\documentclass[10pt,letterpaper]{article}

\usepackage{cogsci}
\usepackage{xcolor}
\usepackage{graphicx}
\usepackage{algpseudocode}
\usepackage{algorithm}
\usepackage{amsmath} 

\usepackage{multirow}
\usepackage{multicol}

\usepackage{url}

\usepackage{breakurl}
\usepackage[breaklinks]{hyperref}

\cogscifinalcopy 

\usepackage{pslatex}
\usepackage{apacite}
\usepackage{float} 

\usepackage{xspace}
\newcommand{\Instruction}[1]{\text{Ins}_{#1}}
\newcommand{\Register}[1]{\text{Reg}_{#1}}
\newcommand{\Symbol}[1]{\text{Sym}_{#1}}
\newcommand{\Answer}[1]{\text{Ans}_{#1}}

\newcommand{\Store}[0]{\texttt{Store}\xspace}
\newcommand{\Ignore}[0]{\texttt{Ignore}\xspace}
\newcommand{\RegisterZero}[0]{\texttt{Reg0}\xspace}
\newcommand{\RegisterOne}[0]{\texttt{Reg1}\xspace}
\newcommand{\Same}[0]{\texttt{same}\xspace}
\newcommand{\Different}[0]{\texttt{different}\xspace}

\title{Transformer Mechanisms Mimic Frontostriatal Gating Operations When Trained on Human Working Memory Tasks}
 
\author{Aaron Traylor,$^1$ Jack Merullo,$^1$ Michael J. Frank,$^2$ \& Ellie Pavlick$^1$ \\
$^1$ Department of Computer Science\\
$^2$ Department of Cognitive, Linguistic \& Psychological Sciences\\
Brown University, Providence, Rhode Island, USA\\
\texttt{\{aaron\_traylor, jack\_merullo, michael\_frank, ellie\_pavlick\}@brown.edu}
}

\begin{document}

\maketitle

\begin{abstract}

Models based on the Transformer neural network architecture have seen success on a wide variety of tasks that appear to require complex ``cognitive branching''-- or the ability to maintain pursuit of one goal while accomplishing others. In cognitive neuroscience, success on such tasks is thought to rely on sophisticated frontostriatal mechanisms for selective \textit{gating}, which enable role-addressable updating-- and later readout-- of information to and from distinct ``addresses'' of memory, in the form of clusters of neurons. However, Transformer models have no such mechanisms intentionally built-in. It is thus an open question how Transformers solve such tasks, and whether the mechanisms that emerge to help them to do so bear any resemblance to the gating mechanisms in the human brain. In this work, we analyze the mechanisms that emerge within a vanilla attention-only Transformer trained on a simple sequence modeling task inspired by a task explicitly designed to study working memory gating in computational cognitive neuroscience. We find that, as a result of training, the self-attention mechanism within the Transformer specializes in a way that mirrors the input and output gating mechanisms which were explicitly incorporated into earlier, more biologically-inspired architectures. These results suggest opportunities for future research on computational similarities between modern AI architectures and models of the human brain.

\end{abstract}

\section{Introduction}

Computational models based on the Transformer architecture \cite{vaswani2017attention} have seen success on a wide variety of tasks that appear to require complex ``cognitive branching'': the ability to maintain pursuit of one over-arching goal while performing other subtasks along the way. For example, Transformer-based large language models (LLMs)  have demonstrated impressive abilities in not just language \cite{brown2020language}, but planning \cite{huang2022language}, navigation \cite{du2021vtnet}, and problem solving \cite{lewkowycz2022solving}.

In humans, performance on such tasks is known to depend on a neural mechanism for \textit{gating} \cite{frank2012mechanisms,badre2012mechanisms,chatham2014corticostriatal,rac2016dissociating,rac2021analogous}, which controls whether new information is maintained in working memory or not, the address in memory where it is stored, and the address from which stored information is recalled in response to a task. 
Typical Transformer models have no specialized architecture for working memory, in spite of their ability to succeed at tasks which appear to require it. Although some Transformer models have additional built-in structure for memory \cite{dai2019transformer, burtsev2020memory}, recurrence \cite{dai2019transformer}, or hierarchy \cite{wang2019tree}, vanilla Transformer models without any such inductive biases remain the dominant architecture for the modern AI systems which are especially visible in their success on complex tasks \cite{brown2020language, touvron2023llama}. This raises the question: in solving such tasks, does a mechanism for selective input and output gating emerge within the vanilla Transformer?

Transformers are good candidates for learning gating behavior because of inductive biases within the self-attention mechanism, i.e., the Transformer's defining architectural component. Within self-attention, numerous ``attention heads'' construct contextual representations for each item in their input sequence through a learned weighted combination of the previous items in the sequence.  Attention heads could in principle learn gating behavior by marking sequence elements with a key (input gating) and reading out those values later by querying for those keys (output gating). Moreover, the attention mechanism in Transformers is decomposed in a way that enables it to readily differentiate ``reading'' and ``writing'' operations. This behavior is analogous functionally to the neurobiological roles of corticostriatal circuits in humans and other animals, in which isolated clusters of prefrontal neurons represent distinct ``addresses'' in memory that can be updated or read out from via selective gating actions triggered by basal ganglia and thalamus \cite{o2006making,frank2012mechanisms,calderon2022thunderstruck}. In computational neuroscience models of this process, the prefrontal clusters (or ``stripes'') can also serve as latent abstract ``roles'' that condition how to interpret content within them, affording functions such as variable binding, indirection, and hierarchical generalization to new situations \cite{o2006making,collins2013cognitive,kriete2013indirection,bhandari2018learning}. Thus, Transformers may use their attention heads to learn a gating strategy that mimics certain functions of the brain.

In this work, we train vanilla Transformer models with self-attention on a working memory task paradigm that was specifically designed to evaluate models of selective gating and working memory in computational neuroscience  \cite{o2006making,rac2021analogous}. We use recent techniques from \textit{mechanistic interpretability} \cite{olahblog,nanda2022transformerlens} to expose the mechanism that the Transformer uses in order to perform the task.
We find that, as a result of training, the self-attention mechanism 
specializes in a way that resembles existing models of input-output gating.
Specifically, we find that the trained model uses the \textit{key vectors} within the attention mechanism to control \textit{input gating}, i.e., determining which elements in the input to consider vs.\ ignore, as well as controlling how the information is stored-- in other words, assigning it a \textit{role} such that the model can access it later. The model uses the \textit{query vectors} to control \textit{output gating}, i.e., determining which information is accessed in order to complete a task. Our findings highlight the importance of considering the emergent mechanisms that result from training in addition to the innate architectural mechanisms when drawing comparisons between AI systems and human cognitive processes, and opens the door for future analysis and work which can enable more principled studies of the similarities and differences between human vs.\ machine cognition.

\section{Background}

\subsection{Gating Mechanisms}

\noindent Working memory in human brains is known to make use of a \textit{gating mechanism}, which processes and stores information roughly analogously to gates being opened and closed \cite{rac2016dissociating,rac2021analogous,bhandari2018learning}.
The input gate controls which information is stored or not stored in memory, and if stored, into which ``address''. The output gate controls which content within working memory is accessed in order to produce a response or to make subsequent gating operations. Both input and output gating policies are also dependent on the learned task-dependent context (i.e.\ \textit{role}) of the information to be stored and accessed in working memory.

In cognitive neuroscience, a variety of tasks are used to study the capacity of working memory. In this work, we focus on a variant of the ``reference-back 2'' task \cite{rac2021analogous}, a human paradigm meant to mimic a task designed to showcase the need for selective gating of independent contents of information in frontostriatal neural networks \cite{o2006making}. In the reference-back paradigm, symbols such as letters or numbers are viewed one at a time, and the subject must determine whether the current symbol is the same or different as that stored in memory for a given role (letter or number). They also are given a cue to indicate whether to update the current symbol as the ``reference'' to be compared on subsequent trials of the same role, or if instead they should continue to maintain the previous reference.  Thus this task requires selective updating and accessing of information in a role-addressable manner. 

\subsection{Transformers}

Transformers are powerful language models which create contextualized representations of sequences of words, where they learn to to predict the next token one at a time using an ``attention'' mechanism \cite{bahdanau2014neural} to scan the previously seen tokens for relevant information. These models are able to learn and represent any arbitrary sequence modelling task.

For a given prediction, Transformer attention generates three separate vectors at each position in the sequence: a query, key, and value ($q$, $k$, $v$). The query vector scans the previous context (including the current token) for relevant keys, and calculates how much the current prediction should ``attend'' to those positions. Then, the value vectors at those positions are multiplied by the corresponding weights, summed up, and added to the next representation: for token $i$ at layer $j$, the contextual representation is $\sum_k q_i^j \cdot k_k^j * v_k^j$. Thus, the next token prediction includes information from earlier in the sequence by combining the value vectors from previous tokens. In other words, Transformer attention can be viewed as a read/write mechanism: for a given token, the queries and keys dictate which tokens to read from, the values are the content that is read proportional to the attention calculated by the keys and queries, and the summed content is written to a new representation at the given token. As we shall see below, the comparison to role-addressable input and output gating operations is evident, whereby the key vectors form addresses analogous to the PFC ``stripes'', with key construction determining input gating, and the query vectors control which of them are accessed, with query construction determining output gating. 
 
\paragraph{Limitations as Cognitive Models:}

Transformers have properties which make them obviously bad models of human sequence processing. In particular, because Transformers can attend to any part of the sequence when creating a representation, they are not limited by memory representation constraints. Transformers could thus solve tasks that would push the limits of human working memory, but it should be noted they accordingly require large amounts of training data. 
The question addressed in this work is orthogonal to these limitations. That is, we 
focus specifically on if and how Transformers can learn to implement an efficient gating mechanism to solve tasks with human working memory demands. 

\begin{figure}[ht!]
    \centering
    \includegraphics[width=\columnwidth]{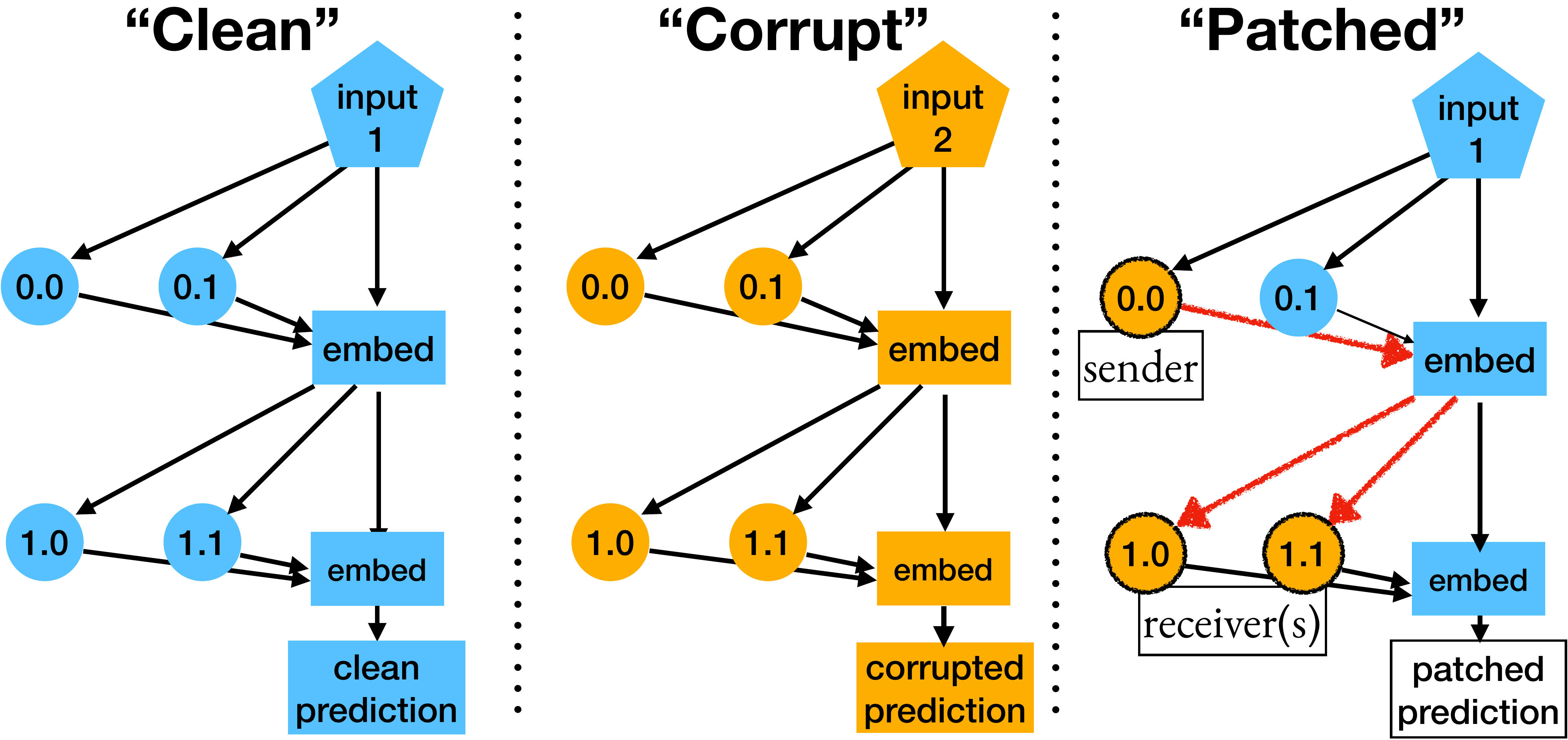}
    \caption{Graphical diagram of the path-patching process. Attention heads are represented as circles (layer,head index), and contextual representations of each token (as well as the next token prediction) are represented as rectangles.}
    \label{fig:pp_example}
\end{figure}

\subsection{Mechanistic Interpretability}
\label{sec:pp}
We use a set of recently introduced analysis tools \cite{elhage2021mathematical} which enable us to uncover specific mechanisms defined in terms of model weights within the Transformer.
Specifically, we use path-patching \cite{wang2022interpretability, goldowsky2023localizing}, a generalization of causal mediation analysis \cite{pearl2001} that allows us to determine which components of a neural model (e.g., attention heads) work together in order to produce observed behavior on a task. The discovered components are referred to as a \textit{circuit} \cite{rauker2023toward}.

Path-patching involves making a incisive edit to the representations of a trained model and observing how the model's behavior is affected (see Fig.\ \ref{fig:pp_example}). Path-patching typically requires a minimal pair of examples: the ``clean'' example and the ``corrupted'' example, in which one token from the clean example is changed, as well as the correct label. Given representations from the model for both the clean and the corrupted examples (the blue and orange components in the figure),
we can chose a specific component anywhere in the model (referred to as the ``sender''), and insert the corrupted representation at that component into the clean representation. We then use the model to recalculate the representations up until another component of the model (the ``receiver''), thus ``patching'' the path. In the figure, we send from layer 0, head 0 to layer 1, both heads 0 and 1. All clean representations that are not along this path are not modified and are unaffected by the patch. The model then recomputes all representations after the receiver (the ``patched'' representations), and arrives at a new prediction. If the model output matches the corrupt prediction rather than the clean one, that prediction is causally dependent on the path from sender to receiver. See \shortcite{wang2022interpretability} and \shortcite{goldowsky2023localizing} for a more comprehensive review of path-patching methods.


\begin{figure}[ht!]
    \centering
    \includegraphics[width=\columnwidth]{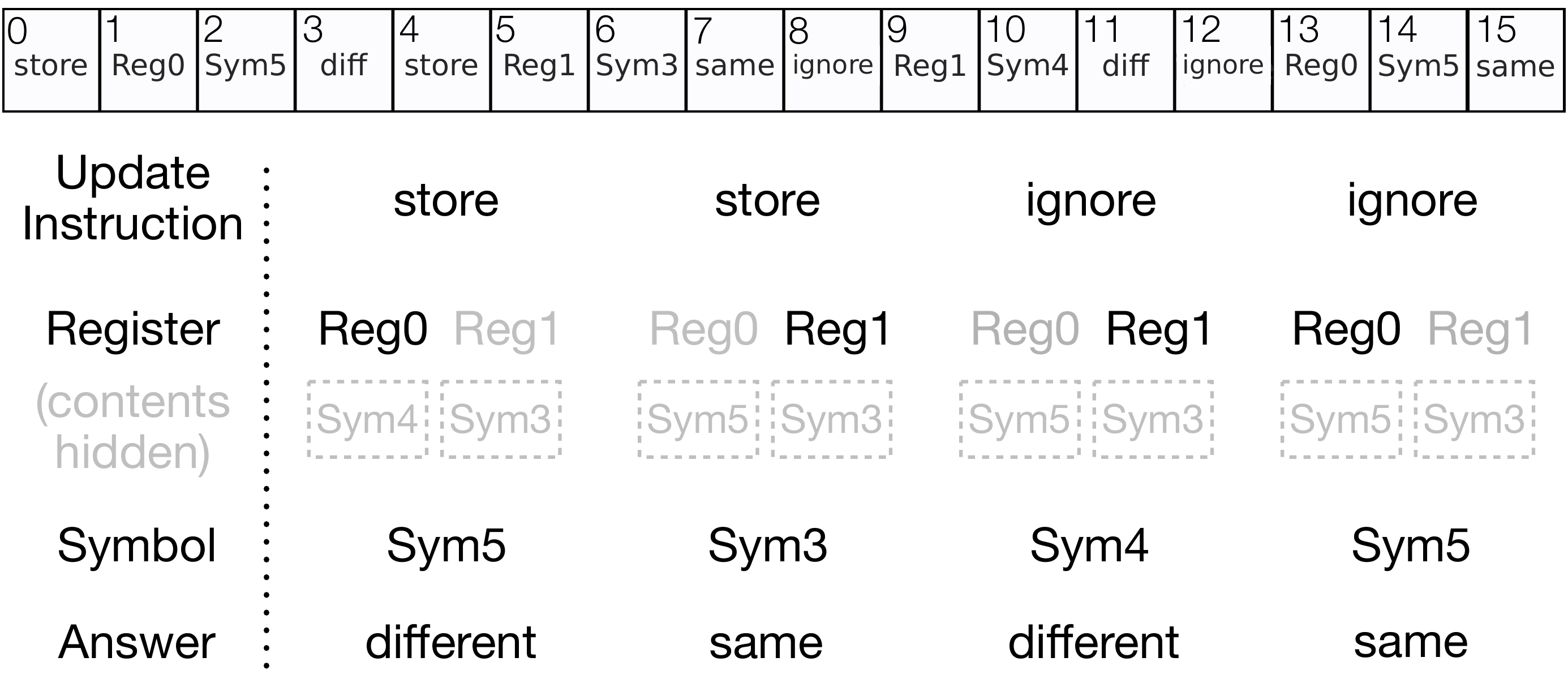}
    \caption{Above: example of textual reference-back task as model input. Below: step-by-step task process; models do not view task-internal grey words. ``Update Instruction'' executes after ``Answer'' despite appearing earlier sequentially.}
    \label{fig:task_example}
\end{figure}

\section{Task}

We create a modified text-based version of the reference-back 2 task \cite{rac2021analogous} designed to tax selective WM gating \cite{o2006making}.
The \textit{textual reference-back task} requires making same/different judgments between incoming symbols assigned to a particular ``register'' in memory, with respect to those seen previously and linked to those same registers. 
Like the original tasks, the textual reference-back task is sequential, and requires the maintenance and independent updating of two memory registers, each containing one of $S$ arbitrary \textit{symbols} at a time. 
The contents of the registers are updated over the course of the task. At the beginning of each sequence, each register is initialized individually to one $s \in S$ (the pool of symbols is shared between registers, which was shown to more substantively tax gating mechanisms in \cite{o2006making}). Each sequence is composed of $L$ tuples, each containing register address $\Register{i}$, symbol $\Symbol{i}$, same/different label $\Answer{i}$, and update instruction $\Instruction{i}$. For a tuple $i \in L$, the \textbf{answer} $\Answer{i}$ is a binary value that is either \Same if symbol $\Symbol{i}$ is currently stored in the register with address $\Register{i}$, or \Different otherwise. The \textbf{update instruction} $\Instruction{i}$ also takes one of two values, evenly distributed: if \texttt{ignore}, then there is no effect further on in the sequence.
If \texttt{store}, then from that point on in the sequence, $\Symbol{i}$ is stored in the register with address $\Register{i}$ until otherwise updated. An example is shown in Fig. \ref{fig:task_example}. 

We implement each reference-back task example in our data as a single sequence, and  measure models' ability to predict \Same versus \Different for each $\Answer{i}$. Each sequence has 10 same/different answers, and we generate 100,000 train, 1,000 dev, and 1,000 held-out test sequences.


The class balance of \Same to \Different answer labels in the train/test datasets is roughly 1:2, making a ``maximum class'' heuristic solution 0.66 accuracy, 0.33 precision, and 0.5 recall. We test several other heuristics, the strongest of which is predicting \Same if another tuple including \Store and the target register and target symbol exists in the sequence, which scores 0.80 accuracy, 0.82 precision, and 0.85 recall.

\section{Model}

We use vanilla Transformers in order to facilitate interpretability, as done in prior work that analyzes emergent mechanisms \cite{elhage2022superposition}. Our models contain two decoder-only layers, each with only two heads (four in total), and no multilayer perceptrons or layer normalization, followed by a linear ``unembed'' layer to project the output of the last decoder into the space of the entire vocabulary at each timestep \footnote{In practice, only `same' and `different' are ever predicted.}. Our network uses absolute positional embeddings \cite{vaswani2017attention}. The vocabulary contains all possible tokens, represented individually with embedding size $E$. Models are trained to predict the next token with the language modelling objective, meaning if the model is predicting $\Answer{c}$, it will have access to all ($\Instruction{i}$, $\Register{i}$, $\Symbol{i}$, $\Answer{i}$) tuples where $i < c$, as well as $\Instruction{c}$, $\Register{c}$, and $\Symbol{c}$. However, the models only receive loss at positions where a same/different token must be predicted. Furthermore, each layer gets a causal attention mask-- when constructing each token representation, it cannot look ahead at tokens further down the sequence. 

The models are trained over 60 epochs of the 100k training data points, learning from 6 million examples in total. Models are evaluated on their accuracy (whether the correct $\Answer{i}$ is predicted for each tuple $i$), measured in precision and recall, as well as the \Same versus \Different token logit difference. 

\section{Experiments}

First, we select and analyze a single Transformer model which succeeds on the task, and upon investigation of its weights find that it learns a mechanism for input/output gating. 
Second, when we conduct a search over more trained models, we find that model performance correlates with markers of learning a gating policy, analogous to findings in the frontostriatal neural networks \cite{frank2012mechanisms}.

\begin{figure*}[ht!]
    \centering
    \includegraphics[width=\textwidth]{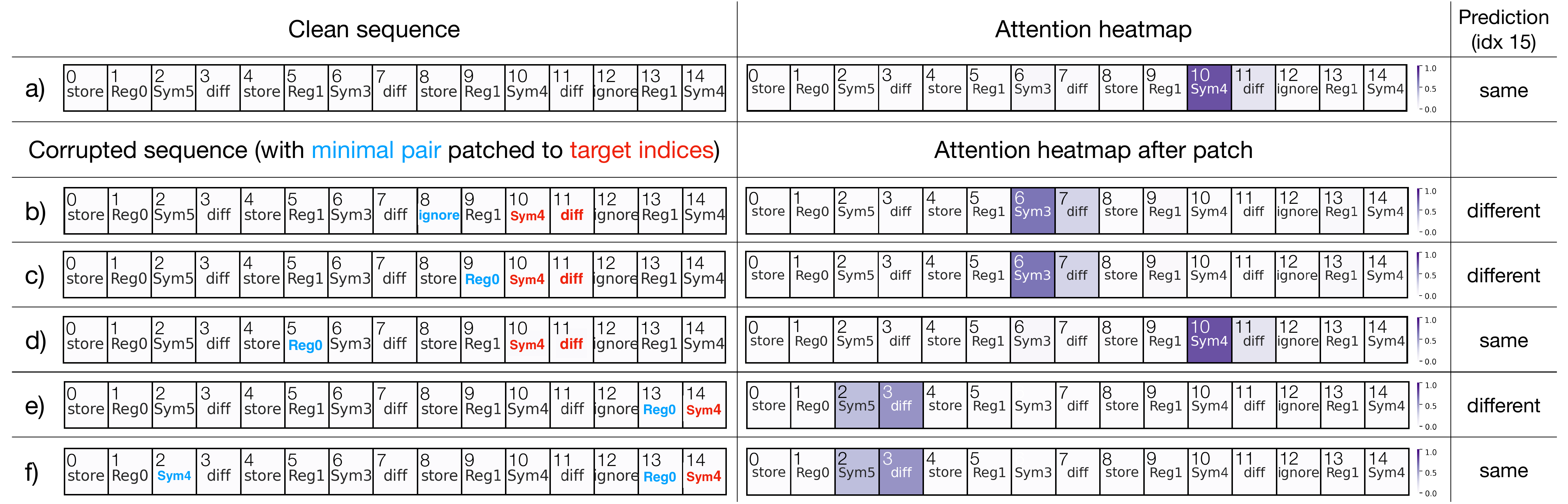}
    \caption{Model behavior when predicting same/different (token 15) is shown. We measure attention visualized as a shade of purple, with deeper shade corresponding to higher attention to that token. We create ``corrupted'' minimal pairs in which changing a token (light blue) either changes the correct label at index 15 (examples b, c, e) or does not (d, f). We make small path-patching edits with the minimal pair to targeted network components (layer 1 keys for b, c, d, f; queries for e,f). In other words, we replace specific components (denoted with red text) with their corresponding representation from the ``corrupted'' sequence, but hold all other representations constant, and run the model and get a new same/different prediction. In all test examples, making the small patch successfully results in the model's prediction changing to align with the ``corrupted'' example.}
    \label{fig:q_and_k_comp}
\end{figure*}

We first establish that a Transformer model is able to succeed on the reference-back task.
We perform a small hyperparameter search 
and select a model that reaches 100\% accuracy on the held-out test data for further analysis. We determine the circuit that the model uses to solve the textual reference-back task through an array of path-patching experiments with a simplistic minimal pairs paradigm.
Our ``sender'' within path-patching is always both attention heads at layer 0, and our ``receiver'' is always both attention heads at layer 1. 

At layer 0, the model learns to condense the task-critical information from each tuple into one embedding, at the position for $\Symbol{i}$\footnote{Redundantly, the model does the same at the position for $\Answer{i}$. Through additional experimentation, we determine that this is a quirk of Transformer learning, and does not impact our analysis.}. At this layer, the model pays 85.8\% of total attention to the task-critical information to that tuple, and just 14.2\% of attention to other tuples. 

At layer 1, the attention heads learn to attend to the $\Symbol{i}$ key vector representing the tuple where information was last stored in the target register. The heads pay 70.2\% of total attention to this tuple (the ``stored'' tuple), and only 29.8\% of attention to all other tokens. This behavior is tied to the target register matching the register in the stored tuple, which is analogous to gating of the relevant role-addressable PFC stripe. We focus our analysis on the Layer 1 representations which exhibit this learned gating policy, shown in Fig. \ref{fig:q_and_k_comp}.



\subsection{Input Gating through Key Vectors}

Input gates in working memory control what incoming information is remembered and ``role-addressable''-- i.e., stored in memory in such a way that it is able to be freely accessed later when it is needed for task completion. In a Transformer, the key vectors serve the role of addresses (analogous to PFC ``stripes''), which are retrieved based on their match to a query vector from the current or later timesteps during self-attention. Thus the input gating in Transformers is controlled through key construction; the composition of the output of the Layer 0 attention controls which content is stored in the key vector at Layer 1 for later use. At a later timestep, a query vector will address the information in the key vector.

In our model analysis, we find that key composition at the $\Symbol{i}$ position (positions 2, 6, 10, and 14 in Fig. \ref{fig:q_and_k_comp}) roughly represents each tuple. A query's ability to address this position depends on whether the represented tuple contains a \Store or an \Ignore. Key vectors representing an \Ignore tuple receive very little attention (0.4\% of layer 1 attention averaged over test set), whereas those representing a \Store tuple receive the bulk of the attention (86.8\%). We determine this effect causally with path-patching (Fig. \ref{fig:pp_example}).
First, we create clean sequences sampled from our test set, and then corrupt these sequences by switching a \Store within tuple $i$ to an \Ignore. We then path-patch only the key vectors of $i$. We expect, if the key controls input gating, that patching these key vectors should ``block'' attention to all of tuple $i$. An example attention pattern is in Fig. \ref{fig:q_and_k_comp}, examples a and b. We find that the model's attention shifts away from the tuple accordingly in 100\% of patched instances.
The presence of an \Ignore or a \Store within a tuple controls whether the key construction acts as an open input gate or a closed input gate. 

Key construction also depends on the \textit{role} of the represented content; within our task, that means whether $\Register{i}$ is \RegisterZero or \RegisterOne. When making a same/different prediction, key vectors representing a tuple that matches the target register receive most of the model's attention (92.5\% of total attention), while those that do not match are not attended to (3.3\% of total attention). Similarly to input gating, we use path-patching to determine that key construction encodes roles. This time, given a target tuple $i$ with a target register, we corrupt the register of the stored tuple, changing it from \RegisterZero to \RegisterOne or vice versa; to predict the answer for $i$, the model must attend to an earlier tuple with the target register. An example is Fig. \ref{fig:q_and_k_comp}, row c. The model's attention shifts away accordingly across every example in the test set. Note that the stored tuple must be modified; if the same corruption is made earlier (as in row d), attention does not shift. This behavior shows that the gating within self-attention is \textit{role-addressable}; the registers within the task function as roles, and are embedded within the key vectors as part of the representation. 



\subsection{Output Gating through Query Vectors}

Within working memory, output gates control which addressable information is accessed in order to complete a task. Given that key vectors serve the role of addresses, query vectors in turn control which key vectors are accessed, through the final Q*K dot product in attention. Query construction thus performs the role of output gating within Transformers.

The query composition controls which addressable $\Symbol{i}$ representations are attended to based on the identity of the target register; changing \RegisterOne to \RegisterZero controls which role-addressable content is accessed. We determine this through a final set of path-patching experiments, an example of which is shown in Fig. \ref{fig:q_and_k_comp}, row e. Rather than editing the register in the stored tuple as done in row c, we corrupt the target register itself; this means that the query must now find representations corresponding to the other tuple. In row e, the model finds \texttt{$Sym_5$\xspace}, and predicts $\Different$; and in row f, we patch in \texttt{$Sym_4$\xspace} at index 2, and the model predicts \Same.
Upon inspection, the query is constructed to correspond with key vectors that represent tuples which also contain the target register. When we edit the target register in minimal pair experiments, we observe that the attention shifts from the original stored tuple (74.1\% of attention) to the stored tuple that matches the edited register, and successfully makes an updated same/different comparison to the symbol in the edited register in every instance. 

Editing aspects of the target tuple other than target register has minimal effect on the query construction behavior. No edits to the query cause the model to attend to a \Ignore tuple, further evidencing of output gating behavior-- only content that has been made ``addressable'' can be accessed for a response. Furthermore, we find that the target instruction and symbol do not factor into the query composition-- changing them through path-patching to the query does not affect attention. This is notable because the model could employ other strategies for determining which tuples are eligible to be the stored tuple; e.g. attending to all symbols to match if any of them are the same as the target symbol.

\subsection{Successful Task Performance is Related to Discovering Gating Policies}

To further identify how readily Transformer models learn a gating policy, and how useful such a policy is to succeed on the task, we train new models with the same hyperparameters across many different random seeds, and measure their performance on the target task as well as on markers of the gating policy.
We train 20 new models on the same textual reference-back data, each with a different random initialization, and measure both training loss and test set accuracy. 5 of the models succeed 100\% of the time, and the other 15 models succeed between 94\%-99.99\% of the time, with a mean of 97.72\% and a standard deviation of 2.03. The models are trained on the same amount of data (6 million examples). 

We observed in the prior sections that the trained Transformer model uses its key composition to control input gating and its query composition to control output gating and role addressability. 
 To identify whether the new models learn to gate similarly, we evaluate the key and query composition of all 20 new models by making minimal pair path-patching edits for every test example, where the answer changes from same to different or vice versa, similarly to Figure \ref{fig:q_and_k_comp}.

To evaluate the ability to open and close input gates, we corrupt the stored tuple's \Store to \Ignore, and path-patch only to the stored tuple's \textit{key vectors}. To evaluate the role addressability of the content during output gating, we corrupt the target register (changing \RegisterZero to \RegisterOne or vice versa), and path-patch only to the \textit{query vector} for making the same/different judgment. This is a more challenging task than patching to all of the keys or queries respectively; a model will only succeed on these subtasks if it implements a gating policy with the same markers that the model analyzed in earlier sections does. A model makes the patched prediction successfully if its prediction matches the corrupted sequence and not the original sequence-- i.e. the targeted path-patch was sufficient to change its same/different prediction.

\begin{figure}[ht!]
    \centering
    \includegraphics[width=\columnwidth]{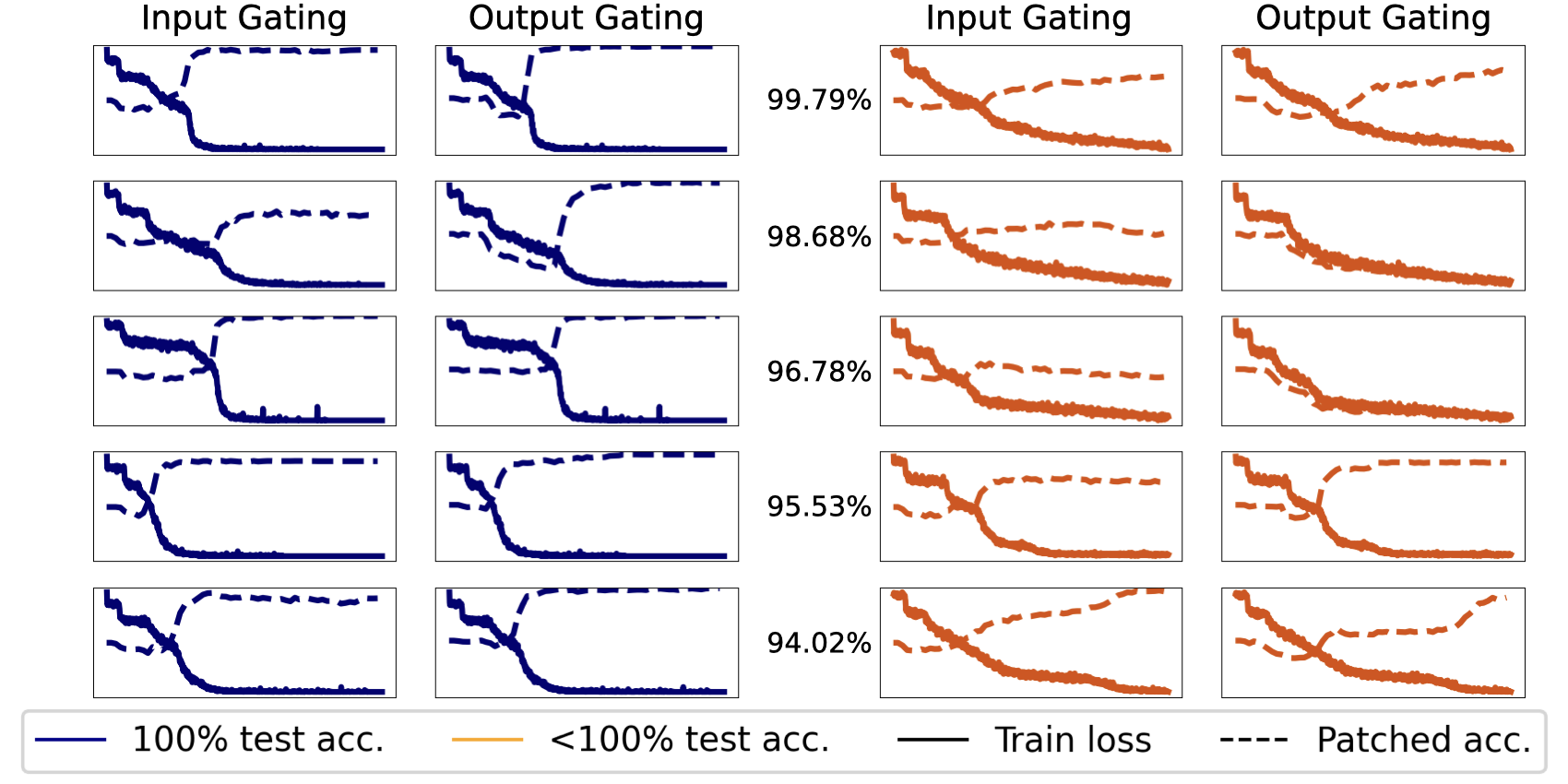}
    \caption{Model performance over training on patching subtasks. Each row corresponds to an individual model's training loss (solid line) and subtask accuracy (dashed line) over time; blue and orange lines respectively correspond to models which reach 100\% test accuracy and to those that do not.}
    \label{fig:expt2}
\end{figure}

We visualize the 5 runs that reach 100\% accuracy on the test data in Figure \ref{fig:expt2}, as well as 5 randomly selected runs that do not reach 100\% accuracy on the test data, and plot their training loss versus their accuracy (between 0 and 100) on the two patched subtasks over the course of training.

Two trends become apparent from the data: first, models which score a perfect test accuracy appear to succeed at the subtasks more readily than models which do not. Of the former, 3 of 5 models reach 100\% accuracy on both subtasks readily, plateauing less than halfway through training duration. However, examples from the latter category of models do not reach such immediate success at the patched subtasks (including the 10 not pictured in this graph); in fact, many categorically fail, scoring as low as 49\% accuracy. These results do not indicate that this class of models' representations are useless for the task-- they all score between 94\% and 99.99\%, well above heuristic performance. Failing to succeed at the targeted patching subtasks reveals that these models may implement some other strategy that is not gating, and may be brittle or heuristic in some way. We take these results as evidence that learning a robust policy for gating correlates with model performance at the textual reference-back task.

The second emergent trend is that many models across both classes have a sharp decline in training loss, which correlates with a similarly steep increase in accuracy on both subtasks.  Sudden jumps in performance is a noted phenomenon that has been observed in other Transformer models in cases of e.g., grokking \cite{power2022grokking}. We interpret this phase transition as suddenly learning a gating mechanism. Models that do not exhibit phase transitions to the same degree take longer to fit the task, and do not reach high subtask accuracy. The cause of phase transitions and what is learned during this process is left for future work.

\section{Summary and Discussion}

In this work, we investigate Transformer models for emergence of a learned \textit{gating mechanism}; a network component performing role-addressable gating, similar to that in working memory of humans. We observe that the model readily learns a gating policy, and upon training more models find that task performance is correlated with gating ability. Our results show how competence at cognitive branching tasks can emerge in Transformers, and suggest that integrating Transformer components may improve existing computational neuroscience models of working memory.

The Transformer models which were trained on the textual reference-back task are capable of making use of key composition for input gating and query composition for output gating. We find that making precise corruptions to specific architectural elements of the network causes the model's prediction to change from \Same to \Different or vice versa, indicating that those components are causally responsible for the gating mechanism. The architectural biases of attention within the vanilla Transformer model lend themselves well to representing role-addressable content, as the learnable nature of keys, queries, and values allows the model to learn to create internal representations in a manner which allows it to represent roles and addresses, mimicking the variable binding and input / output gating mechanisms in biological neural networks \cite{o2006making,frank2012mechanisms,collins2013cognitive,kriete2013indirection}.

In this work, we focused on characterizing the mechanism that the Transformer model learns as a result of training on cognitive branching tasks, and did not evaluate the robustness of the mechanism. The textual reference-back task is similar in nature to the FFLM task \cite{liu2023exposing}, comprised of 1 register, 2 symbols (0 or 1), and long sequences. \citeauthor{liu2023exposing} found that similar Transformer models were able to succeed on FFLM data, but struggled to generalize outside of their training distribution. We leave experiments characterizing generalization of Transformer mechanism behavior on data with the reference-back paradigm-- e.g. more than 2 registers, distinct sets of symbols, novel symbols introduced at test time-- for future work.

When we trained more models on the task, we found that the models which perform best on the task correlate with the markers of gating we observed in our circuit analysis, and that the learning trajectory shows a steep decrease in training loss and a steep rise in patched subtask accuracy simultaneously, suggesting that the model has learned a component of the gating policy at that time. Both findings are analogous to those of \citeA{frank2012mechanisms}, in which they find that networks which learned a hierarchical gating policy performed better at a hierarchical learning task, and humans that learn this policy also show a sharp decrease in loss when they discover it. 

Ultimately, finding connections between emergent behavior of Transformer models and human working memory serves to benefit both computational cognitive neuroscience and artificial intelligence. Although Transformer models themselves are limited in their biological plausibility, in this setting they learned behavior mimicking the functionality of working memory, and their application within computational models of the brain should be further explored. From the perspective of artificial intelligence, understanding the strengths and limitations of Transformer models on cognitive branching tasks may inform model analysis across the many diverse settings in which these models are applied.


\bibliographystyle{apacite}

\setlength{\bibleftmargin}{.125in}
\setlength{\bibindent}{-\bibleftmargin}

\bibliography{bibliography}

\end{document}